\documentclass[10pt,twocolumn,letterpaper]{article}

\usepackage{iccv}
\usepackage{times}
\usepackage{epsfig}
\usepackage{graphicx}
\usepackage{amsmath}
\usepackage{amssymb}

\usepackage{comment}
\usepackage{ifthen} 
\usepackage{overpic}
\usepackage{booktabs}
\usepackage{multirow}
\usepackage{rotating}
\usepackage{amsmath}
\usepackage{amssymb}
\usepackage{caption}
\usepackage{pifont}
\usepackage{colortbl} 
\usepackage{color}
\usepackage{marvosym}
\usepackage{mdframed}
\usepackage{float}
\usepackage{textcomp}
\usepackage{float}
\usepackage[pagebackref=true,breaklinks=true,letterpaper=true,colorlinks,bookmarks=false]{hyperref}

\iccvfinalcopy 

\def\iccvPaperID{12999} 

\ificcvfinal\fi

\def\eg{\emph{e.g.}}

\begin{document}

\title{Layton: Latent Consistency Tokenizer for 1024-pixel Image \\ Reconstruction and Generation by 256 Tokens}

\author{Qingsong Xie\\
OPPO AI Center\\
{\tt\small xieqingsong1@oppo.com}
\and
Zhao Zhang\\
Nankai University\\
{\tt\small zzhang@mail.nankai.edu.cn}
\and
Zhe Huang\\
Tsinghua University\\
{\tt\small huangz23@mails.tsinghua.edu.cn}
\and 
Yanhao Zhang\\
OPPO AI Center\\
{\tt\small zhangyanhao@oppo.com}
\and 
Haonan Lu\\
OPPO AI Center\\
{\tt\small luhaonan@oppo.com}
\and 
Zhenyu Yang\\
OPPO AI Center\\
{\tt\small yangzhenyu@oppo.com}
}

\makeatletter
\g@addto@macro\@maketitle{
    \vspace{-30pt}
        \begin{figure}[H]
          \setlength{\linewidth}{\textwidth}
          \setlength{\hsize}{\textwidth}    
          \centering
          \includegraphics[width=0.956\linewidth]{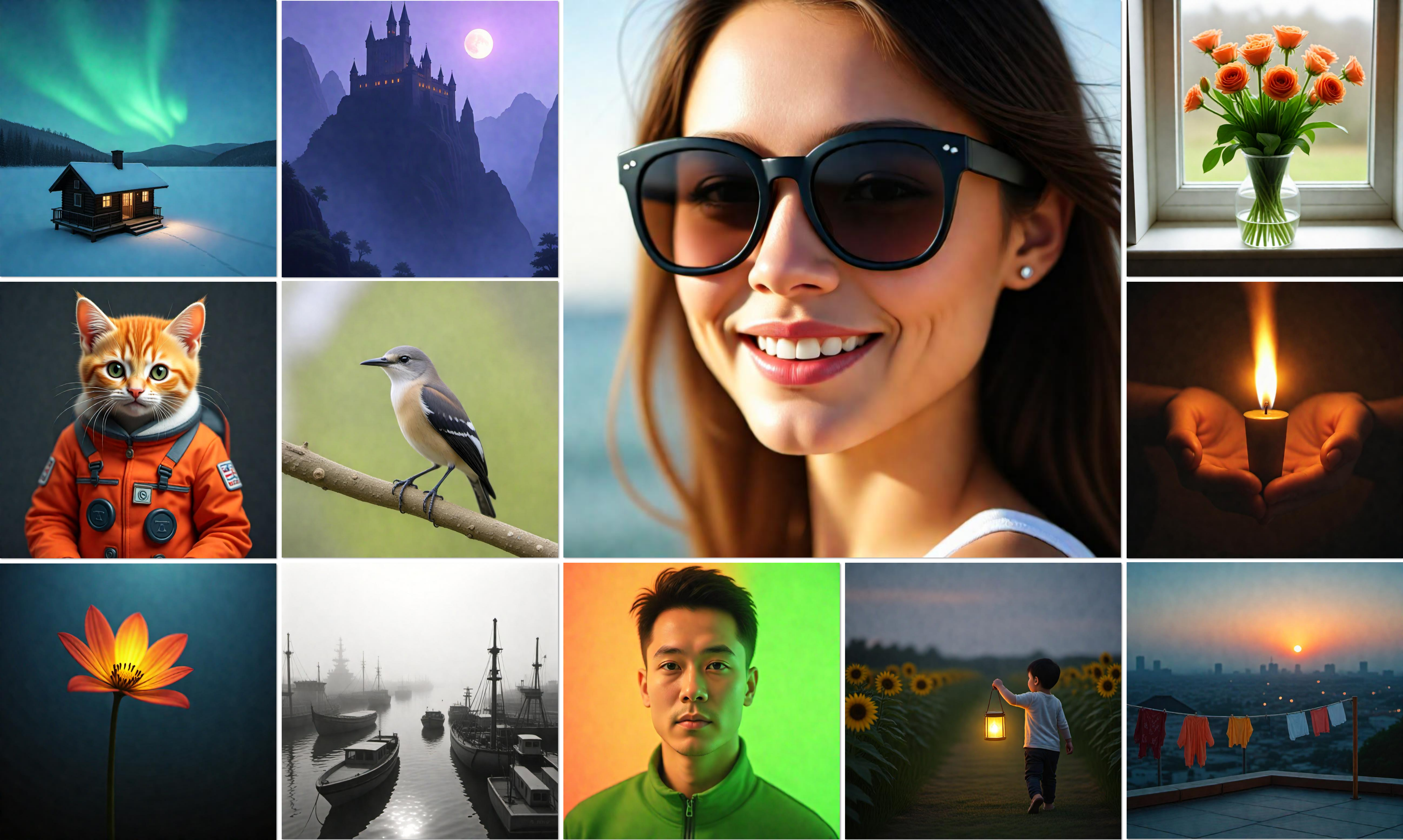}
          \caption{1024-pixel  image generation results of LaytonGen-T* in an autoregressive way with \textbf{256 tokens}, demonstrating the capability of  LaytonGen-T* in high-quality image synthesis.
          }
        \end{figure}
}
\makeatother

\maketitle
\ificcvfinal\thispagestyle{empty}\fi

\begin{abstract}
Image tokenization has significantly advanced visual generation and multimodal modeling, particularly when paired with autoregressive models. However, current methods face challenges in balancing efficiency and fidelity: high-resolution image reconstruction either requires an excessive number of tokens or compromises critical details through token reduction. 
To resolve this, we propose Latent Consistency Tokenizer (Layton) that bridges discrete visual tokens with the compact latent space of pre-trained Latent Diffusion Models (LDMs), enabling efficient representation of 1024×1024 images using only 256 tokens—a 16$\times$ compression over VQGAN. 
Layton integrates a transformer encoder, a quantized codebook, and a latent consistency decoder.
Direct application of LDM as the decoder results in color and brightness discrepancies;
thus, we convert it to latent consistency decoder, reducing multi-step sampling to 1-2 steps for direct pixel-level supervision.
Experiments demonstrate Layton’s superiority in high-fidelity reconstruction, with 10.8 reconstruction Frechet Inception Distance on MSCOCO-2017 5K benchmark for 1024×1024 image reconstruction.
We also extend Layton to a text-to-image generation model, LaytonGen, working in autoregression.
It achieves 0.73 score on GenEval benchmark, surpassing current state-of-the-art methods.
Project homepage: https://github.com/
OPPO-Mente-Lab/Layton

\end{abstract}

\section{Introduction}

\label{sec:intro}
Image tokenizer~\cite{van2017vqvae1,yuimage} aims to convert images from their raw pixel-based representations into discrete visual tokens, which can then be used to reconstruct the original image through its corresponding decoder.
This approach has garnered significant attention due to its crucial role in image generation, particularly in autoregressive models~\cite{sun2024llamagen,tian2024var,wang2024emu3} and masked transformers~\cite{chang2023muse,chang2022maskgit,ding2022cogview2,xie2024show}.

A representative approach, VQGAN~\cite{yu2021vitvqgan} learns a codebook to quantize continuous embeddings of a 256$\times$256 image into 256 discrete tokens using a spatial downsampling ratio of 16$\times$—a standard configuration in recent works~\cite{mizrahi20244m,qu2024tokenflow,shi2022divae,sun2024llamagen}.
When it comes to high-resolution image reconstruction or generation, \eg, 1024$\times$1024 pixels, requires predicting 4096 tokens, creating substantial challenges in both computational efficiency and model optimization.
This lengthy token sequence complicates downstream applications, such as integrating visual tokens into large multimodal models for interleaved text and image understanding and generation.
While recent advancements explore token compression through residual codebooks ~\cite{lee2022rqvae,tian2024var} and 1D tokenization~\cite{kim2025democratizing,yuimage} for 256$\times$256 images, these approaches have not substantially addressed the challenges of higher-resolution image reconstruction and generation.
Recent latent diffusion models (LDMs)~\cite{esser2024sd3,flux2023,podell2023sdxl} have demonstrated remarkable success in 1024$\times$1024 image generation by operating in low-dimensional latent spaces. This raises a compelling question: Can discrete visual tokens be aligned with the compact latent space of LDMs to leverage their powerful decoders for high-fidelity reconstruction and generation use? 

In this paper, we introduce a \underline{La}tent Consistenc\underline{y} \underline{To}ke\underline{n}izer  (Layton), comprising a transformer encoder, a quantized vector codebook, and a latent consistency decoder. 
Our key insight is to align discrete visual tokens with the compact latent space of pre-trained LDMs, enabling efficient representation of 1024$\times$1024 images with only 256 tokens—a 16$\times$ reduction compared to VQGAN.  Inspired by ControlNet~\cite{zhang2023adding}, we first employ a latent diffusion decoder through copying adaptive blocks in LDM with zero convolution connecting copy and raw LDM.
During training, Layton is optimized using diffusion objectives with progressive resolution scaling from 512$\times$512 to 1024$\times$1024. However, relying solely on diffusion loss results in reconstructed images with noticeable discrepancies in color and brightness. To address this, we introduce latent consistency models~\cite{ren2025hyper,xie2024tlcm} to the decoder, converting the multi-step sampling process into one or two steps. It enables more direct supervision and improving reconstruction fidelity.
Moreover, we extend Layton to text-to-image generation model (LaytonGen) by training an autoregressive transformer, which  efficiently generates these compact token sequences through text-instructed autoregressive prediction. 

We conducted comprehensive experiments to evaluate Layton on the ImageNet 50K~\cite{deng2009imagenet},  MSCOCO-2017 5K~\cite{lin2014microsoft} benchmarks. With 256 tokens to reconstruct 1024$\times$1024 pixel images, Layton achieves  10.80 reconstrution Frechet Inception Distance (rFID) score~\cite{heusel2017gans}  on MSCOCO-2017, significantly outperforming VQGAN~\cite{yu2021vitvqgan}, TiTok~\cite{yuimage},  LlmaGen~\cite{sun2024llamagen}. On ImageNet benchmark, Layton gets 2.78 rFID, surpassing VQGAN and LlamaGen, comparable to TiTok, but with much better Peak Signal-to-Noise Ratio, Structural Similarity Index Measure, and Learned Perceptual Image Patch Similarity than TiTok. For text-to-image task, LaytonGen  obtains 0.73 score on GenEval benchmark~\cite{ghosh2023geneval},  superior to other diffusion and autoregressive models, e.g., 0.62 by SD3~\cite{esser2024scaling}, 0.56 by HART~\cite{tanghart}, 0.53 by Show-o~\cite{xie2024show}, and 0.32 by LlamaGen.

Our contributions are threefold:
\begin{itemize}
    \item We propose Layton, an image tokenizer that bridges discrete visual tokens with the latent space of pre-trained LDMs, demonstrating the ability to reconstruct and generate 1024$\times$1024 images with only 256 tokens.
    \item To alleviate color and brightness discrepancies caused by diffusion objective in reconstruction, we integrate latent consistency model into latent diffusion decoder  with direct pixel-level supervision through efficient few-step sampling.
    \item  Built on Layton, LaytonGen can generate high-quality 1024$\times$1024 images via text-guided autoregressive token prediction. It achieves leading performances on multiple synthesis benchmarks.
\end{itemize}

\section{Related Work}
\label{sec:related_work}
\subsection{Image Tokenization}
Variational Autoencoders 
(VAEs)~\cite{kingma2013vae} represent a significant advancement in the field by learning to map inputs to a distribution. Building upon this foundation, VQVAEs~\cite{van2017vqvae1} learn discrete representations that form a categorical distribution. This process is further improved in VQGAN~\cite{esser2021taming}, which enhances the training process through the integration of adversarial training techniques.
The transformer architecture within autoencoders is explored in ViT-VQGAN~\cite{yu2021vitvqgan}. RQ-VAE~\cite{lee2022rqvae}, introduces residual quantization to the VAE framework, recursively quantizes the feature map in a coarse-to-fine manner, allowing for a precise approximation of the feature map with a fixed codebook size.
In a different vein, MAGVIT-v2~\cite{yu2023MAGVIT-v2}, FSQ~\cite{mentzerfinite}, BSQ-ViT~\cite{zhao2024image} propose lookup-free quantization, presenting an alternative approach that bypasses traditional lookup mechanisms. 
TiTok~\cite{yuimage} performs 2D-to-1D distillation, compressing the number of tokens used to represent the same image.
VILA-U~\cite{wu2024vila}, TokenFlow~\cite{qu2024tokenflow}, and TA-TiTok~\cite{kim2025democratizing} introduce text supervision to enhance semantic information for discrete tokens.
DiVAE~\cite{shi2022divae} introduces a diffusion decoder, turning single-step decoding into a multistep probabilistic process at the pixel level. However, they can not reconstruct or generate high-resolution image details by small number of tokens. Motivated by LDMs~\cite{esser2024scaling,flux2023,podell2023sdxl} which are able to generate high-resolution, high-quality images, we explore effective method to leverage the pre-trained LDM as decoder to reconstruct high-resolution images. 

\subsection{Tokenized Image Generation}
Image tokenization has become a powerful technique for image generation, allowing images to be represented as discrete tokens that can be manipulated and generated using various modeling approaches. Two prominent methodologies in this domain are the masked-transformer style and the autoregressive style.
In the masked-transformer style, methods like MaskGIT~\cite{chang2022maskgit} employ a bidirectional transformer decoder. During training, the model predicts randomly masked tokens and iteratively refines the image at inference. Other notable works in this category include~\cite{chang2023muse,lee2022draft,lezama2022improved,lezama2023predictor}.
Conversely, the autoregressive style involves predicting all tokens of an image in a serialized manner, akin to next-word prediction~\cite{esser2021taming,lu2022unified}. 
 This approach naturally integrates with the multimodal large language model~\cite{jin2023unified,team2024chameleon,wu2024vila,zhan2024anygpt} and is conducive to unifying various modalities.
LlamaGen~\cite{sun2024llamagen} stands out as a simple yet effective generation method that adopts the autoregressive approach, but it fails to generate satisfactory images. We extend Layton to conditional generation model by training an autoregressive model to predict the next discrete tokens produced by the tokenizer. Thanks to the excellent performance of Layton to reconstruct images, our model is able to generate superior images. 

\section{Preliminary}
\label{sec:preliminary}
\subsection{VQGAN.}
VQGAN~\cite{esser2021taming}  typically consists of  encoder $Enc$, quantizer $Q$, and decoder $Dec$.  Given an image $\mathbf{x} \in \,\mathbb{R}^{H\times W\times 3}$, where $H,W$ denotes image height and width, respectively,  $Enc$ firsts extracts its latent embeddings $\mathbf{G}=Enc(\mathbf{x}) \in \mathbb{R}^{H/f\times W/f\times D}$,  effectively
reducing the spatial dimensions by a factor of $f$. Then,  $Q$ maps each embedding $\mathbf{g}\in \mathbb{R}^D$ in $\mathbf{G}$ with $D$ representing embedding dimension to the nearest
code $c_i$ in a learnable codebook $\mathbb{C}\in \mathbb{R}^{N \times D}$, where $N$ is the codebook size. Mathematically, this can 
be formulated as: 
\begin{equation}\label{eq:vq}
Q(\mathbf{g}) = \mathbf{c}_{Tok}, \ \ Tok = \arg\min_{j \in \{1, 2, \dots, N\}} \|\mathbf{g} - \mathbf{c}_j\|_2^2.
\end{equation}
The mapped feature vectors  $\mathbf{C} \in \mathbb{R}^{H/f\times W/f\times D}$ are calculated by $Q(\mathbf{G})$, while  
decoder converts  $\mathbf{C}$  to  image  $\hat{\mathbf{x}}$  through $Dec(\mathbf{C}))$.

\subsection{Diffusion Models.}
 Diffusion model~\cite{ho2020denoising} is composed of a forward diffusion process and a reverse denoising process. Forward process gradually adds random noise to clean data $\mathbf{x}_0$ and diffuses it into pure Gaussian noise as:
 \begin{equation}\label{eq:diff}
     \mathbf{x}_t=\alpha_t \mathbf{x}_0+ \beta_t\mathbf{\epsilon}, \quad t\in [0, T],
 \end{equation}
 where $\mathbf{\epsilon} \in \mathcal{N}(0,\mathbf{I}) $, $\alpha_t$, $\beta_{t}$ are scheduler coefficients with $\alpha_t^2+\beta_t^2=1$, and $T$ is the ending time. 
 During the training stage, DMs usually minimizes diffusion loss $\mathcal{L}_{DF}$  via neural network $\epsilon_{\theta}$ predicting noise as:
 \begin{equation}
      \mathcal{L}_{DF}=\| \epsilon_{\theta}(\mathbf{x}_t,t)-\mathbf{\epsilon} \|_2^2.
 \end{equation}
Denoising process reverses the diffusion
process to create clean samples from the Gaussian noise. 

 To reduce resource consumption, Latent Diffusion Models (LDMs)~\cite{esser2024sd3,flux2023,podell2023sdxl} transform pixel image $\mathbf{x}_0$ into latent embedding $\mathbf{z}_0$ via VAE, and perform diffusion and denoising operations in latent space. Sequentially, VAE decoder converts predicted $\mathbf{z}_0$ back to the pixel image. LDMs has shown great success in text-to-image, image-to-image tasks, but they have not been explored for building discrete tokenizer. In this study, we  investigate effective method which enables LDMs to construct discrete tokenizer for high-resolution image reconstruction and generation. 
 

\begin{figure*}[ht]
\vspace{-3ex}
  \centering
  \includegraphics[width=1\linewidth]{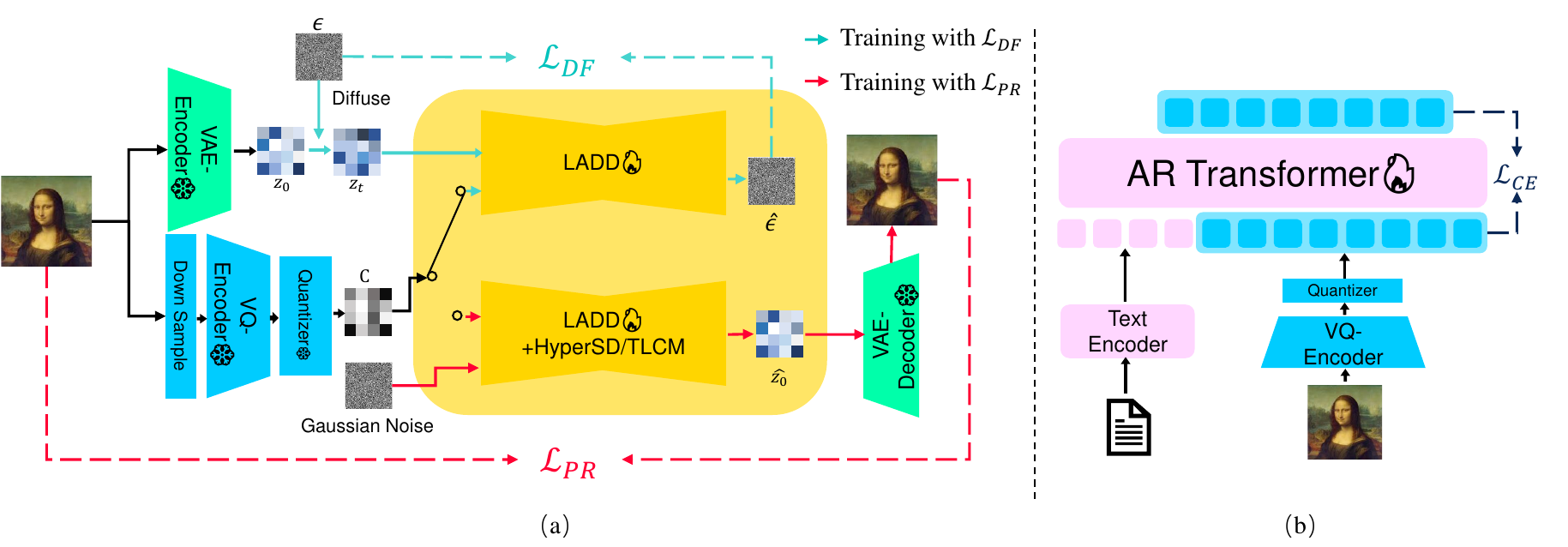}
  \caption{(a)Overview of Layton. 
  The input image is sequentially processed by downsampler, encoder and quantizer into condition features. 
 It also goes through the  VAE Encoder in pretrianed LDM to produce latent $z_0$, which will be diffused to produce $z_t$. LADD takes $C$, $z_t$ and $t$ as input. In the first phase, we apply $\mathcal{L}_{DF}$ to train LADD. In the second phase, we introduce acceleration models to LADD and replace $z_t$ with Gaussian noise to perform one or two step inference, which allows us to train LADD with pixel reconstruction loss $\mathcal{L}_{PR}$.  For simplicity, we omit the time step $t$. (b) Illustration of autoregressive text-to-image generation with LaytonGen.
  }\label{fig:sys}
\end{figure*}

 \section{Methodology} 
 Conventional discrete tokenizers~\cite{esser2021taming,sun2024llamagen,van2017vqvae1,yu2021vitvqgan,yuimage} encode images into discrete tokens and decode the tokens into original images in pixel space. These works  transform text-to-image generation into next-prediction task, which predict image tokens
by autoregressive model  conditioned on the text input. However, these tokenizers need a large number of tokens to generate high-resolution images, leading to prohibitive training and inference costs. Moreover,    these tokenizers can not faithfully reconstruct images with intricate, high-frequency details such as human faces.  Such drawback sets a low ceiling for text-to-image generation task. 

As a result, we are targeting at constructing an innovative tokenizer for high-resolution image reconstruction and generation via a small number of discrete tokens. To this end, we propose Latent Consistency Tokenizer (Layton) which is capable of reconstructing and producing high-quality images with only 256 tokens. Unlike the existing tokenizers that operate image encoding and decoding   in pixel space, Layton models decoding procedure in latent space through unleashing the potential of  latent diffusion model, which further makes use of latent consistency models~\cite{ren2025hyper,xie2024tlcm}, to enforce pixel reconstruction. Upon Layton, we build an autoregressive model for high-resolution image generation driven by text conditions.  Fig. \ref{fig:sys} illustrates the overview of the proposed method. 

\subsection{Latent Diffusion Reconstruction}

Considering that  latent diffusion models~\cite{esser2024scaling,podell2023sdxl} are powerful to synthesize high-quality images, we propose latent diffusion decoder (LADD), denoted as $f_\theta$,  to build discrete tokenizer. The core thought of LADD is to predict the latent representation  $\mathbf{z}_0$ conditioned on the  pixel image $\mathbf{x}_0$. To fulfill this idea, 
the raw image $\mathbf{x}_0 \in \mathbb{R}^{H\times W\times 3}$ is converted into  latent code $\mathbf{z}_0 \in \mathbb{R}^{H/8\times W/8 \times C} $ using VAE encoder of pre-trained LDM, where $H=W \in \{512,1024\}$. 
To predict $\mathbf{z}_0$, diffusion loss is used to train LADD as:
\begin{equation}
\mathcal{L}_{DF}   = \Vert f_\theta(\mathbf{z}_t,\mathbf{C}, t)-\mathbf{\epsilon} \Vert_2^2, 
\end{equation}
where $\mathbf{z}_t$ is obtained by forward diffusion procedure using Eq. (\ref{eq:diff}) in latent space. The condition $\mathbf{C}$ is obtained by sequentially encoding $\mathbf{x}_0$ using $Enc$ and the quantizer $Q$ through the Eq. (\ref{eq:vq}). To reduce token number, $\mathbf{x}_0$ is first downsampled into the resolution of $H'\times W'$, where $H'=W'\in \{224,256,288\}$. 
To promote training's stability, we draw inspiration from ControlNet~\cite{zhang2023adding} to design  LADD structure. Specifically, 
LADD freezes the parameters of pre-trained LDM and
simultaneously  clones some blocks of the  LDM to a trainable copy. Zero convolution ($ZC$) is utilized to connect trainable copy and raw LDM.  The output $\mathbf{O}$ of LADD block at timestep $t$ is computed as:
\begin{equation}\label{eq:LADD}
    \mathbf{O}=ZC(F_{train}(z_t,\mathbf{C},t))+ F(z_t,t),
\end{equation}
where $F$ and $F_{train}$ denote  the frozen LDM block and    trainable block in  trainable copy, respectively.

To train our tokenizer and save training resource, the visual encoder and quantilizer are initialized  using pre-trained LlamaGen tokenizer. Next, we freeze the visual encoder and quantilizer, enabling only LADD to undergo training. 

\begin{figure*}[ht]
  \centering
  \includegraphics[width=0.9\linewidth]{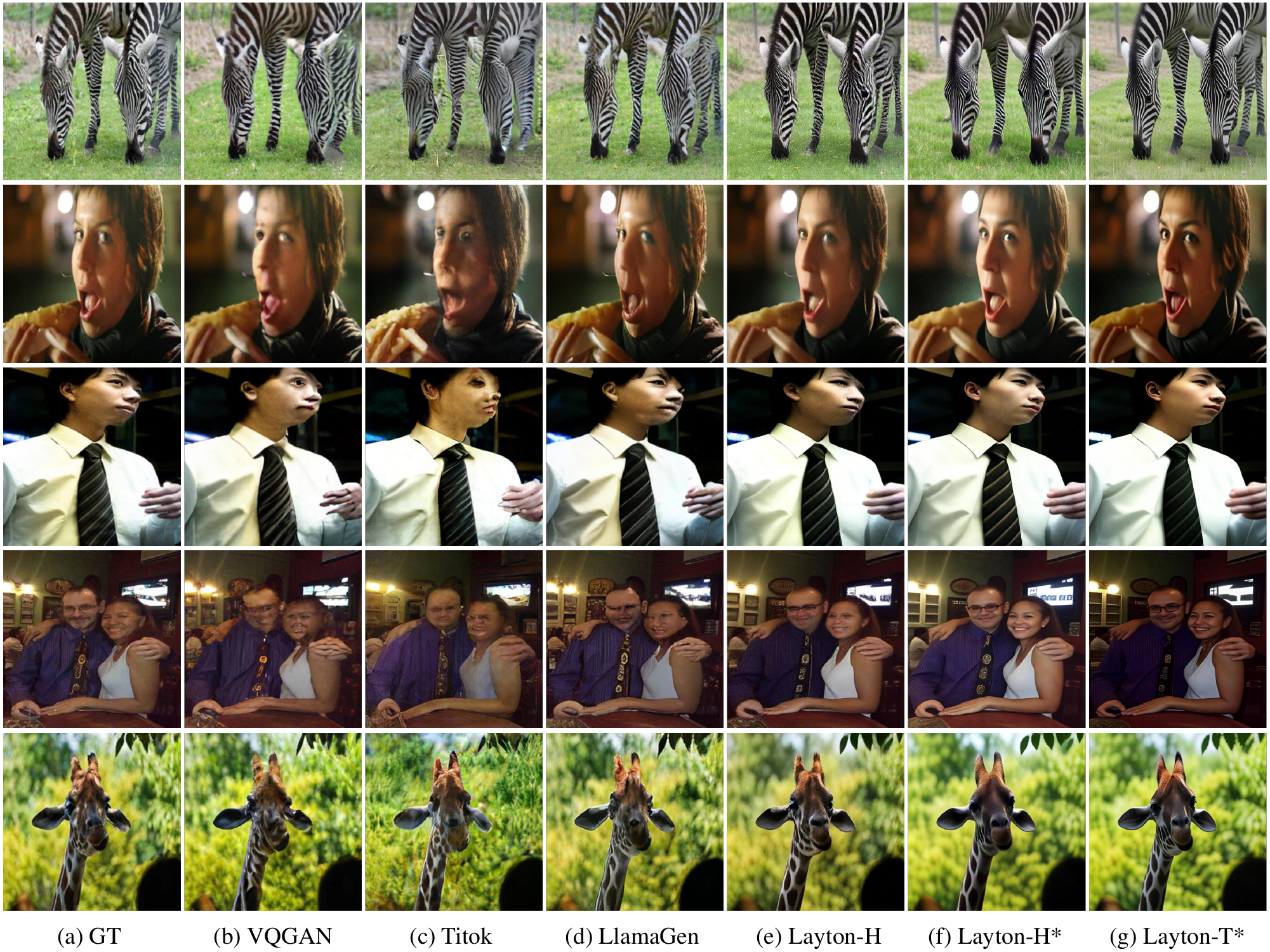}
  \caption{Visual comparisons of images  reconstruction for different methods. 
 Layton can achieve much better reconstruction results than VQGAN, TiTok, and LlamaGen, especially in facial reconstruction. The images reconstructed by Layton-H* and Layton-T* show higher quality than other methods, even surpass ground truth (GT).  
  }\label{fig:rec}
\end{figure*}

\subsection{Pixel Reconstruction}

The diffusion loss enables the decoder to reconstruct the original image through multi-step sampling. However, we have empirically observed that the  reconstructed images  exhibit discrepancies in color and brightness compared to the original images. To address this issue, we introduce a pixel reconstruction loss $\mathcal{L}_{PR}$ that compels the decoded image to recover the original image. Since  diffusion model requires multi-step sampling to generate images, it is unable to  directly minimize $\mathcal{L}_{PR}$, which  consumes a significant amount of GPU memory and may lead to gradient explosion. Constency model (CM) can enable LDMs to generate images with few-step inference. Therefore,  we  leverage CM to  assist LADD in achieving rapid image reconstruction. As Hyper-SD~\cite{ren2025hyper} and Training-efficient Latent Consistency Model (TLCM)~\cite{xie2024tlcm} show state-of-the-art performance, both of which are CMs for LDM'acceleration, they are integrated into LADD to reconstruct image with few steps. Sequentially,  $\mathcal{L}_{PR}$  can be used to optimize our tokenizer as gradient is easily propagated into decoder. 
 One-step sampling is used to reconstruct clean latent code $\mathbf{\hat{z}}_0$ when HyperSD is merged into LADD, because it can generate high-quality image with one step. Two-step sampling is leveraged to restore $\mathbf{\hat{z}}_0$  when incorporating TLCM into LADD  since it needs at least two steps, where stop-gradient operation is adopted for the first iteration. Both models accept pure Gaussian noise as initial latent code.  The reconstruction loss $\mathcal{L}_{PR}$ is:
\begin{equation}
  \mathcal{L}_{PR}=\mathcal{L}_{P}(Dec(\mathbf{\hat{z}}_0), \mathbf{x}_0)  
\end{equation}
where $Dec(.)$ represents the pre-trained VAE decoder in LDM, $\mathcal{L}_P$ is a perceptual loss from LPIPS~\cite{zhang2018unreasonable}. We use Layton-H and Layton-T to represent that Hyper-SD and TLCM are utilized in LADD to minimize $\mathcal{L}_{PR}$, respectively.  


\begin{figure*}[ht]
  \centering
  \includegraphics[width=1\linewidth]{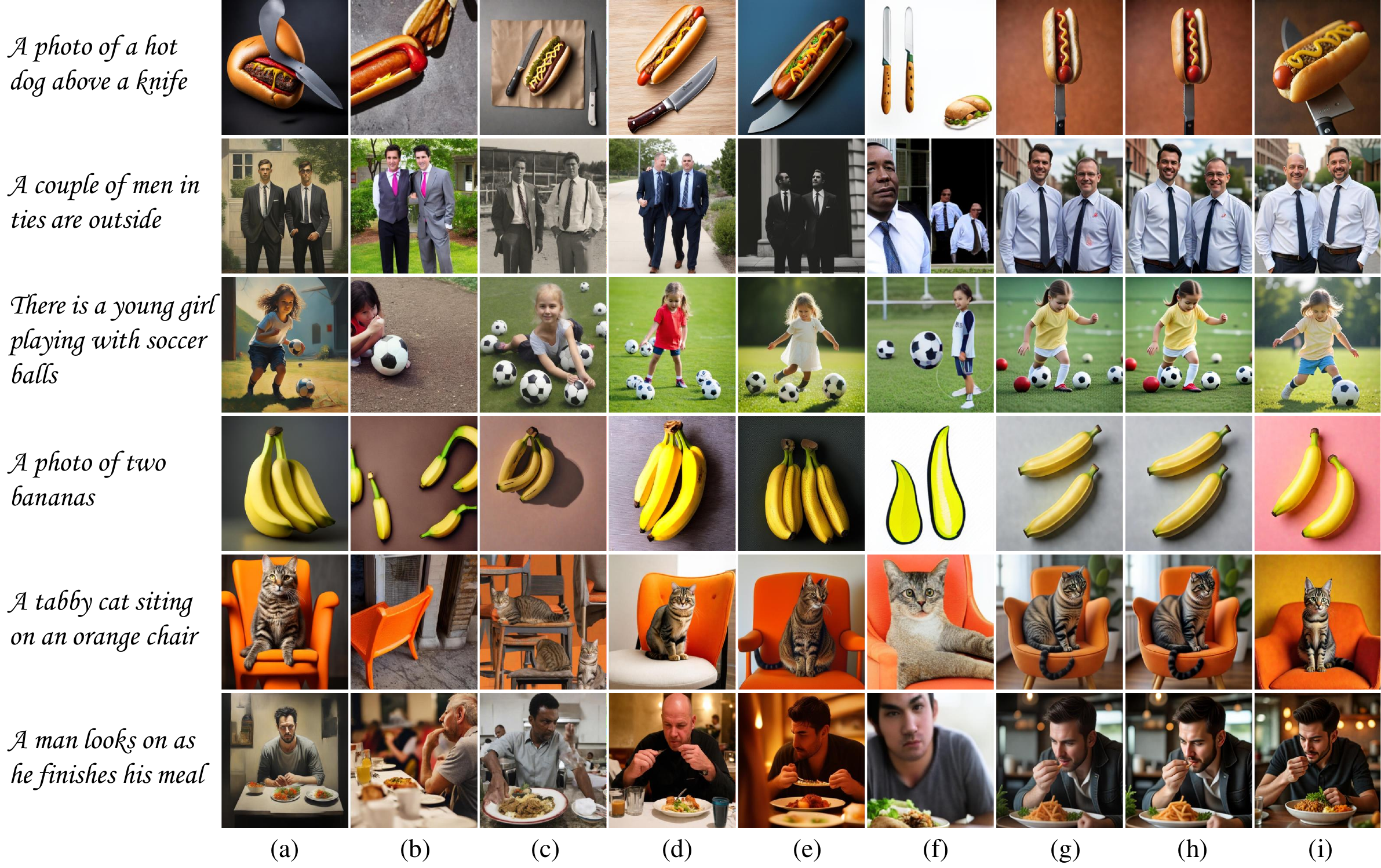}
  \caption{Comparison of text-conditioned generation of different methods. 
  From left to right, (a)HART\cite{tang2024hart}, (b)SD1.5\cite{rombach2022high}, (c)SDXL\cite{podell2023sdxl}, (d)SD3\cite{esser2024sd3}, (e)Show-o\cite{xie2024show}, (f)LlamaGen\cite{sun2024llamagen}, (g)LaytonGen-H, (h)LaytonGen-H* and (i)LaytonGen-T*.
  Apart from satisfactory visual quality, Layton can also yield improved metrics compared to strong baselines. 
  }\label{fig:t2i}
\end{figure*}

\subsection{Text-to-image Generation }
In order to unleash the  value of tokenizer, 
We leverage Layton to build text-to-image generation model (LaytonGen). LaytonGen is implementaed through autoregressive models $P_\theta$ with $\theta$ denoting parameters. A text encoder is used to    extract text feature $\mathbf{f}_{text}$, which  is projected by an
additional MLP to match the dimension of autoregressive models.  
  Cross entropy loss $\mathcal{L}_{CE}$ is applied to train autoregressive model as:
\begin{equation}
    \mathcal{L}_{CE}=-\sum_{i=1}^L\log P_{\theta}(Tok_{i+1}|Tok_{i:1},\mathbf{f}_{text}),
\end{equation}
where $L$ is token number to represent a image. As classifier-free guidance (CFG)~\cite{ho2022classifier} is critical to generate high-quality image in LDM, we adopt
it in our models. During training, the conditional is randomly replaced by a null
unconditional embedding.
During inference stage,  the logit \( \ell_g \) is computed as \( \ell_g = \ell_u + s(\ell_c - \ell_u) \) for every token, where \( \ell_c \) represents the conditional logit, \( \ell_u \) denotes the unconditional logit, and \( s \) is the scaling factor for CFG.

\subsection{Data Construction}
It is well known that high-quality data is necessary to train LDM, but it is hard to access such real data. To deal with this challenge, we use synthesized data instead of real data to train LaytonGen.  After comparing the generated image quality of all open-source models, we select FLUX.1-dev~\cite{flux2023} to synthesize data  as it presents the highest human preference. The text input for FLUX.1-dev is from LAION-5B~\cite{schuhmann2022laion}. Since raw prompt is too simple to describe image content, we use Qwen2-VL~\cite{wang2024qwen2} to re-caption the generated images. As
Some  generated images are not satisfactory, ImageReward (IR)~\cite{xu2023imagereward}  and Multi-dimensional Preference Score (MPS)~\cite{zhang2024learning}  are used to filter undesirable images, where the images with IR $<$ 0.8 or MPS $<$ 12.0 are removed. Totally, we synthesize 40M high-quality text-image pairs.  



\begin{table*}[t]\small
\centering
\caption{The reconstruction performance of Layton on ImageNet, MSCOCO-2017 5K validation dataset, and MJHQ-5K.  *Trained on synthesized data by FLux.1-dev.  }\label{tab:rec}
\begin{tabular}{lccccccccccccc}
        \toprule
    	\multirow{2}{*}{Methods} &\multirow{2}{*}{\#Tokens} & \multicolumn{4}{c}{ImageNet} & \multicolumn{4}{c}{MSCOCO-2017} &\multicolumn{4}{c}{MJHQ-5K} \\
        \cmidrule(l){3-6}\cmidrule(l){7-10}\cmidrule(l){11-14}
        & & rFID$\downarrow$ & P$\uparrow$ & S$\uparrow$ & L$\downarrow$ & rFID$\downarrow$ & P$\uparrow$ &S$\uparrow$& L$\downarrow$ &rFID$\downarrow$ & P$\uparrow$ &S$\uparrow$& L$\downarrow$ \\
        \midrule
        VQGAN~\cite{esser2021taming}&256&6.02&19.20&0.60&0.44&14.72&18.78&0.6&0.46&17.58&18.46&0.64&0.45\\
TiTok-S-128~\cite{yuimage}&128 &\textbf{2.32}&16.97&0.51&0.49      &12.31&16.47&0.50&0.51&                14.17&16.42&0.54&0.50 \\
LlamaGen~\cite{sun2024llamagen}&256&  3.17&\textbf{19.94}&0.64&0.41&11.23&\textbf{19.53}&\textbf{0.64}&0.43&13.26&\textbf{19.24}&0.68&0.41 \\
Layton-H(Ours)&256&2.78&19.80&\textbf{0.65}&\textbf{0.39}
&\textbf{10.80}&19.28&\textbf{0.64}&\textbf{0.41}&\textbf{11.34}&19.16&0.68&\textbf{0.38} \\
Layton-H*(Ours)&256&9.04&19.02&\textbf{0.65}&0.43
&16.93&18.20&\textbf{0.64}&0.43&14.02&18.04&0.68&0.39 \\
Layton-T*(Ours)&256&8.71&18.80&\textbf{0.65}&0.43
&15.05&17.98&\textbf{0.64}&0.42&13.32&17.81&\textbf{0.69}&\textbf{0.38} \\
\hline
\end{tabular}
\end{table*}
\section{Experimental Results}

\subsection{Image Reconstruction}
We employ  Peak Signal-to-Noise Ratio (P), Structural Similarity Index Measure (S) to evaluate the image similarity between reconstructed and raw images. Concurrently, we evaluate the distribution discrepancy   by   reconstruction Frechet Inception Distance score (rFID)~\cite{heusel2017gans}.  Learned Perceptual Image Patch Similarity (L)~\cite{zhang2018unreasonable} is further used to measure perceptual similarity as it  performs excellently in terms of human visual perception. The validation is conducted on ImageNet
50K validation set~\cite{deng2009imagenet},  MSCOCO-2017 5K validation set~\cite{lin2014microsoft}, MJHQ-5K validation set which is randomly sampled from MJHQ-30K dataset~\cite{mjhq2023}. 
All the  images are resized to $1024\times 1024$ size for evaluation as we aim at high-resolution image reconstruction. 

The performance of different tokenizers is listed in Tab.~\ref{tab:rec}. It can be observed that on ImageNet, TiTok-S-128~\cite{yuimage} gets the slightly lower rFID, but much worse P, S, and L values than Layton-H. This result indicates that
compared to Layton-H, the data distribution of images reconstructed by TiTok is closer to that of the original images. However, the reconstructed images by Layton-H exhibit higher image quality and better structural consistency with the original images, particularly aligning more closely with human visual perception.  On MSCOCO-2017  and MJHQ-5K validation sets, Layton-H obtains significant gains than TiTok with respect to all metrics, which demonstrates that Layton-H presents considerably superior generalization performance.  We can also find Layton-H  outperforms LlamaGen~\cite{sun2024llamagen} in terms of rFID, S, L metrics, while achieving comparable PSNR value. This is because  Layton-H takes advantage of pre-trained LDM assisted by HyperSD to  reconstruct images, which is  more powerful to restore image details, e.g., human face.  
Moreover, rFID of  Layton-H* and  trained on synthesized data is  higher than Layton-H and  LLamaGen, because data distribution of synthesized data dramatically
 deviates from those of ImageNet, MSCOCO, and MJHQ sets. 
 The performance of Layton-H* concerning rFID, L are improved by Layton-T*, thanks to its ability to reconstruct intricate, high-frequency by increasing the  sampling steps. 

 Fig. \ref{fig:rec} illustrates the visual comparisons of image reconstruction for different tokeniers.  We can observe that the proposed Layton-H significantly improves reconstruction performance than VQGAN, TiTok, and LlamaGen, especially in image details such as faces. Besides, it  can be seen that the images reconstructed by Layton* and Layton-T* show better quality than other methods, even outperforms ground truth (GT), which implies the potential of our tokenizer to implement super-resolution task. The result also indicates
 that lower rFID on Imagenet, COCO, MJHQ  does not necessarily indicate better reconstruction with respect to human preference.  

\begin{table*}[t]\small
\centering
\caption{The performance of our text-to-image models   on GenEval and MSCOCO-2017 5K validation dataset. }\label{tab:t2i}
\begin{tabular}{lccccccccc}
        \toprule
    	\multirow{2}{*}{Methods} & \multicolumn{5}{c}{GenEval$\uparrow$} & \multicolumn{4}{c}{MSCOCO-2017$\uparrow$} \\\cmidrule(l){2-6}\cmidrule(l){7-10}
         & Two Obj &Counting & Position & Color Attri. & \textbf{Overall} & IR  &MPS& HPSv2 &FS \\
    	\midrule
            \multicolumn{10}{l}{Diffusion Models} \\
            \midrule
           
            SD1.5~\cite{rombach2022high}   & 0.38 &0.35 & 0.04 & 0.06 & 0.43 & 0.16& 10.08&0.285&0.70 \\
            SDXL~\cite{podell2023sdxl} & \textbf{0.74} &0.39 & 0.15 & 0.23 &  0.55 & 0.82& 11.9&0.295 & 2.94 \\
            SD3 ~\cite{esser2024sd3}  & \textbf{0.74} &0.63 & 0.34 & 0.36 & 0.62 & \textbf{1.00}&\textbf{12.59}&0.303&\textbf{3.74}\\
            \midrule
            \multicolumn{9}{l}{AutoRegressive Models} \\
            \midrule
            LlamaGen~\cite{sun2024llamagen}  & 0.34 &0.21& 0.07 & 0.04 & 0.32 & 0.29&9.56&0.273&0.53\\
            HART~\cite{tang2024hart}  & - & - & -& - & 0.56 & 0.66&11.69&0.298 &3.50\\
            Show-o~\cite{xie2024show}  & 0.52 & 0.49& 0.11 &  0.28 & 0.53 & 0.95&10.58&0.277&2.08 \\
            LaytonGen-H&0.67&0.80&0.49&0.50&0.71
            &0.86&12.22&0.298&2.96 \\
            LaytonGen-H*&0.69&0.82&0.51&0.51&0.72 &0.88&12.30&0.302&3.20 \\
            LaytonGen-T*&0.72&\textbf{0.84} &\textbf{0.53}&\textbf{0.51}&\textbf{0.73}&0.90&12.38&\textbf{0.304}&3.68 \\
            \hline
\end{tabular}
\end{table*}

\subsection{Text-to-image Generation}
To evaluate the conditional generation performance of Layton, we perform large-scale text-to-image experiments using LaytonGen. The evaluation is conducted on GenEval benchmark~\cite{ghosh2023geneval} and MSCOCO-2017 5K validation dataset~\cite{lin2014microsoft}. GenEval benchmark is used to  evaluate compositional image properties, such as spatial
relations and attribute binding. 
On MSCOCO dataset,
ImageReward (IR),  Multi-dimensional Preference Score (MPS), and HPSv2~\cite{wu2023human} are used to assess  human preference of the  generated image, while FaceScore (FS)~\cite{liao2024facescore} is adopted to  measure image's face quality. 

As listed in Tab. \ref{tab:t2i}, LaytonGen-T*  remarkably improves the overall metric of LlamaGen  on the GenEval benchmark by 0.41. 
On MSCOCO-2017 5K validation dataset, LaytonGen-T* again outperforms LlamaGen by a large margin in terms of all metrics. These results indicate that the images generated by LaytonGen-T* significantly outperform those from LlamaGen in terms of overall quality and facial quality. The reason lies in LaytonGen-T*'s ability to fully utilize the generative capabilities of the pre-trained LDM, enabling it to produce intricate, high-frequency  details. 
When compared to SDXL~\cite{podell2023sdxl}, LaytonGen-T* improves the metrics by 0.18 on the Geneval benchmark, while on MSCOCO-2017  dataset, all metrics are enhanced again.  One reason behind this is that our Layton-T* can effectively describe intricate,  high-resolution images with small number of discrete tokens, allowing LaytonGen-T* to only train an autoregressive model to predict these discrete tokens, which is easier than directly training SDXL through diffusion loss. Another reason is that we have constructed high-quality data to train the autoregressive model.
It can also be observed that LaytonGen-T* significantly surpasses other autoregressive models, Show-o and HART on Geneval and MSCOCO-2017 dataset, because they reconstruct images  via simple CNN or transformer decoder, failing to reconstruct image details, spatial relations or attribute binding by discrete tokens.  Excitingly,  on GenEval benchmark, LaytonGen-T* beats  SD3~\cite{esser2024sd3} with superior structure MMDiT than UNet in SDXL, and     exhibit competitive generation performance on MSCOCO-2017 dataset, which implies that LaytonGen-T maybe improved by adopting SD3 as the  decoder. Compared to LaytonGen-H, LaytonGen-H* achieves higher  image quality, which results from that the synthesized data facilitates the tokenizer in generating better images. The performance of  LaytonGen-H* is further improved by LaytonGen-T*, which again verifies that TLCM is more powerful than HyperSD to help decoder generate high-quality images. 

As shown in Fig. \ref{fig:t2i}, we can observe that the images generated by LaytonGen-T* enjoy better text-image alignment than other methods, and higher human preference than SD1.5~\cite{Rombach_2022_CVPR}, SDXL, HART~\cite{tang2024hart}, Show-o~\cite{xie2024show}, LlamaGen,  especially in facial generation. 

\begin{table}[t]\small
\label{tab:compare}
\centering
\caption{Ablation study of key components in  Layton.
VQ-LADD means substituting decoder in LlamaGen with the proposed LADD, and using 25-step DDIM to reconstruct images. }\label{tab:ablation}
\begin{tabular}{lcccc}
        \toprule	Methods&rFID$\downarrow$&P$\uparrow$&S$\uparrow$ &L$\downarrow$ \\
        \midrule
        
        LlamaGen&13.26&\textbf{19.24}&\textbf{0.68}&0.41\\
        VQ-LADD&12.72&16.53&0.63&0.47 \\
        VQ-LADD+Hyper-SD&14.71&16.64&0.64&0.47 \\
         VQ-LADD+TLCM&14.40&16.82&0.65&0.45 \\
         Layton-H&\textbf{11.34}&19.16&\textbf{0.68}&\textbf{0.38} \\
\hline

\end{tabular}
\end{table}

\begin{table}[t]\small
\centering
\caption{The effect of token number for Layton-H to  reconstruct 1024-pixel image on MJHQ-5K dataset.}\label{tab:token}
\begin{tabular}{lcccc}
        \toprule	\#Tokens&rFID$\downarrow$&P$\uparrow$&S$\uparrow$ &L$\downarrow$ \\
        \midrule
        192&12.35&18.46&0.66&0.40\\
        256&11.34&19.16&0.68&0.38 \\
        324&\textbf{11.01}&\textbf{19.33}&\textbf{0.69}&\textbf{0.37} \\
\hline
\end{tabular}
\end{table}

\begin{table}[t]\small
\centering
\caption{The effect of CFG scale for LaytonGen-T* to  generate image on MSCOCO-2017 5K validation set.}\label{tab:cfg}
\begin{tabular}{lcccc}
        \toprule	scale&IR$\uparrow$&MPS$\uparrow$&HPSv2$\uparrow$ &FS$\uparrow$ \\
        \midrule
        1.5&0.86&12.37&0.303&3.59\\
        2&0.89&12.38&\textbf{0.304}&3.68\\
        3&\textbf{0.90}&12.40&\textbf{0.304}&3.70 \\
        7&\textbf{0.90}&\textbf{12.41}&\textbf{0.304}&\textbf{3.80} \\
\hline
\end{tabular}
\end{table}


\subsection{Ablation Study}
As shown in Tab. \ref{tab:ablation}, we conduct several experiments to verify the effectiveness of the  key components with respect to  our Layton, where LlamaGen is adopted as baseline. 

\textbf{Latent diffusion decoder.}
Compared with LLamaGen, LADD shows worse metrics concerning P, S, L. The reason lies in the fact that LADD predicts the latent representation of the raw image, leading to discrepancies in color and lighting compared to the original image. 

\textbf{LADD acceleration.}
LADD+Hyper-SD and LADD+TLCM denote that one-step Hyper-SD and three-step TLCM are directly used to accelerate reconstruction procedure. It can be seen that TLCM outperforms Hyper-SD. The probable reason is that TLCM is stronger  to reconstruct the fine details of the image due to its better generation performance. 

\textbf{Reconstruction loss.} Through  optimizing VQ-LADD using pixel reconstruction loss, Layton-H surpasses LLamaGen and VQ-LADD by a large margin in terms of rFID, S, and L metrics. The result indicates that the proposed  reconstruction loss is critical to improve tokenizer's performance, which is able to enforce the reconstructed image close to raw image in pixel space. 

\textbf{Token number.} As summarized in Tab. \ref{tab:token}, we can see even using 192 tokens, Layton-H can reconstruct 1024-pixel image with high performance.  As the increase of the token number, the performance is further improved. 

\textbf{CFG scale.} Tab. \ref{tab:cfg} lists the performance  of  Layton-T* for text-to-image generation with different CFG scale. It can be observed that Layton-T* is capable of generating high-quality image using different CFG. The performance can be improved with higher CFG, and the improvement becomes slight when scale$>$2.

 \section{Conclusion} 
We present Layton, a discrete tokenizer, which  leverages the pretrainded LDMs assisted by acceleration models for 1024-pixel image reconstruction with only 256 tokens. It is also extended to text-to-image generation models, LaytonGen, through an autoregressive model. Extensive experiments demonstrate our Layton outperforms the existing methods for high-resolution reconstruction. Our LaytonGen shows strong capability to generate high-quality images, achieving 0.73 score on GenEval benchmark.
 



{\small
\bibliographystyle{ieee_fullname}
\bibliography{main}
}
\clearpage
\setcounter{page}{1}
\maketitlesupplementary

\section{Implementation Details}
We adopt  the same encoder structure in  LLamaGen and SDXL as the pre-trained LDM.  For tokenizer training, diffusion loss is first used to train Layton for 80000 iterations, and then reconstruction loss is exploited to further optimize Layton for 30000 iterations. 
4 A100  is used with   Adam Optimizer to train Layton, where     $\beta_{1} = 0.9$, $\beta_2 = 0.99$, learning rate (lr) = 1e-5, total  batch size = 4. All the images $\mathbf{x}_0$  with progressive resolution from 512 to 1024 pixel size are  encoded into latent space by SDXL-VAE. The images  are also randomly resized to $\{224,256,288\}$ pixel size, feeding into encoder and quantilizer, yielding  condition of  LDM. During the training stage of LaytonGen, flan-t5-xl is utilized to  extract text features, GPT-XL architecture with 675M parameters in LlamaGen is used as AR model. In this stage, we set lr=1e-4, batch size=320 with 8 A100. AdamW optimizer with $\beta_{1} = 0.9$, $\beta_2 = 0.99$ is used to train LaytonGen for 200K iterations. All the tokenizers are trained on ImageNet, except for those specifically marked with an asterisk (*), which are trained on synthetic data.

\section{Qualitative Visualization Results}
\begin{figure*}[ht]
  \centering
  \includegraphics[width=0.95\linewidth]{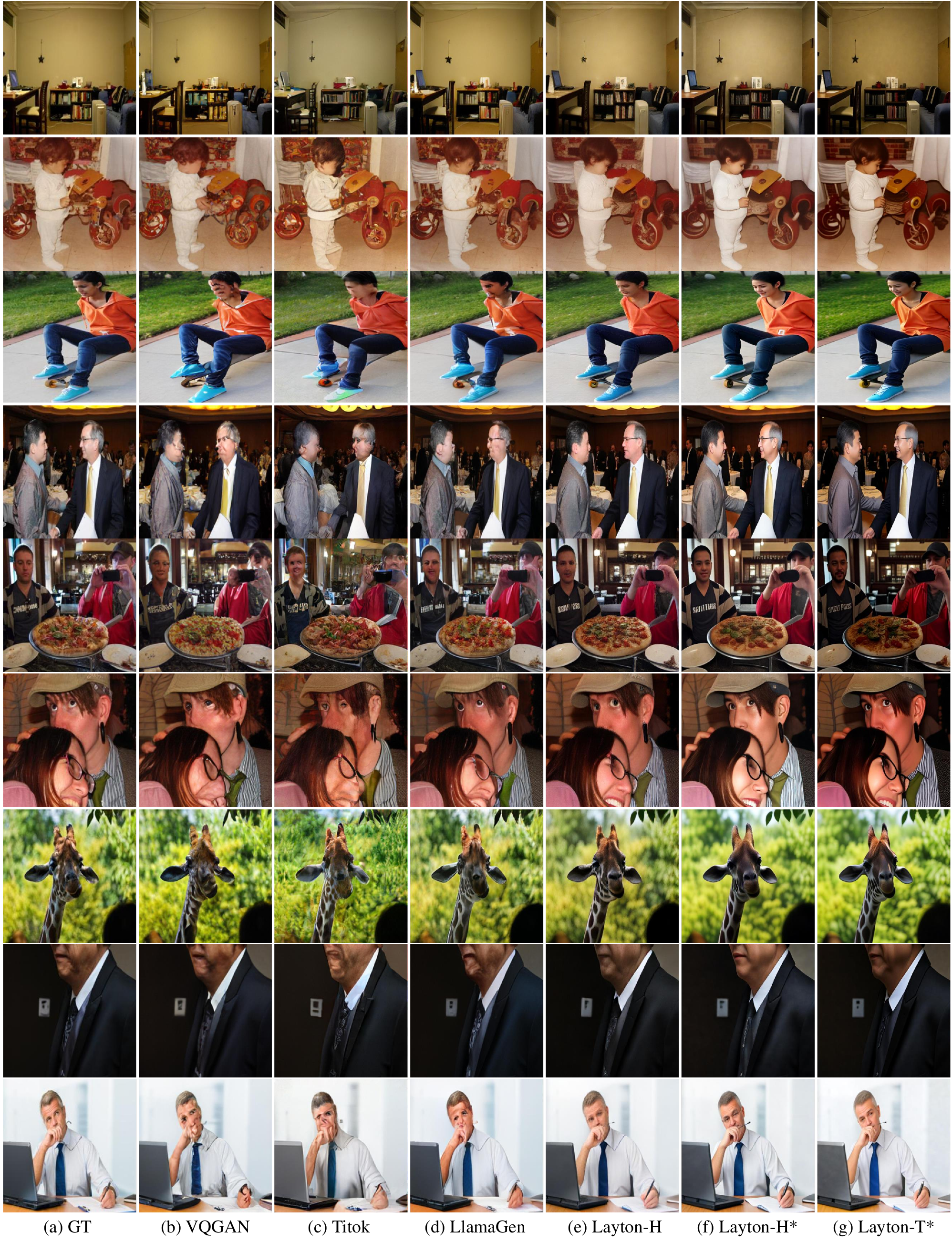}
  \caption{More examples on visual comparisons of images  reconstruction with different methods. 
 From left to right, ground truth, VQGAN, TiTok, LLamaGen, Layton-H, Layton-H* and Layton-T*.}
  \label{fig:reconstruction_appendix}
\end{figure*}

\begin{figure*}[ht]
  \centering
  \includegraphics[width=0.95\linewidth]{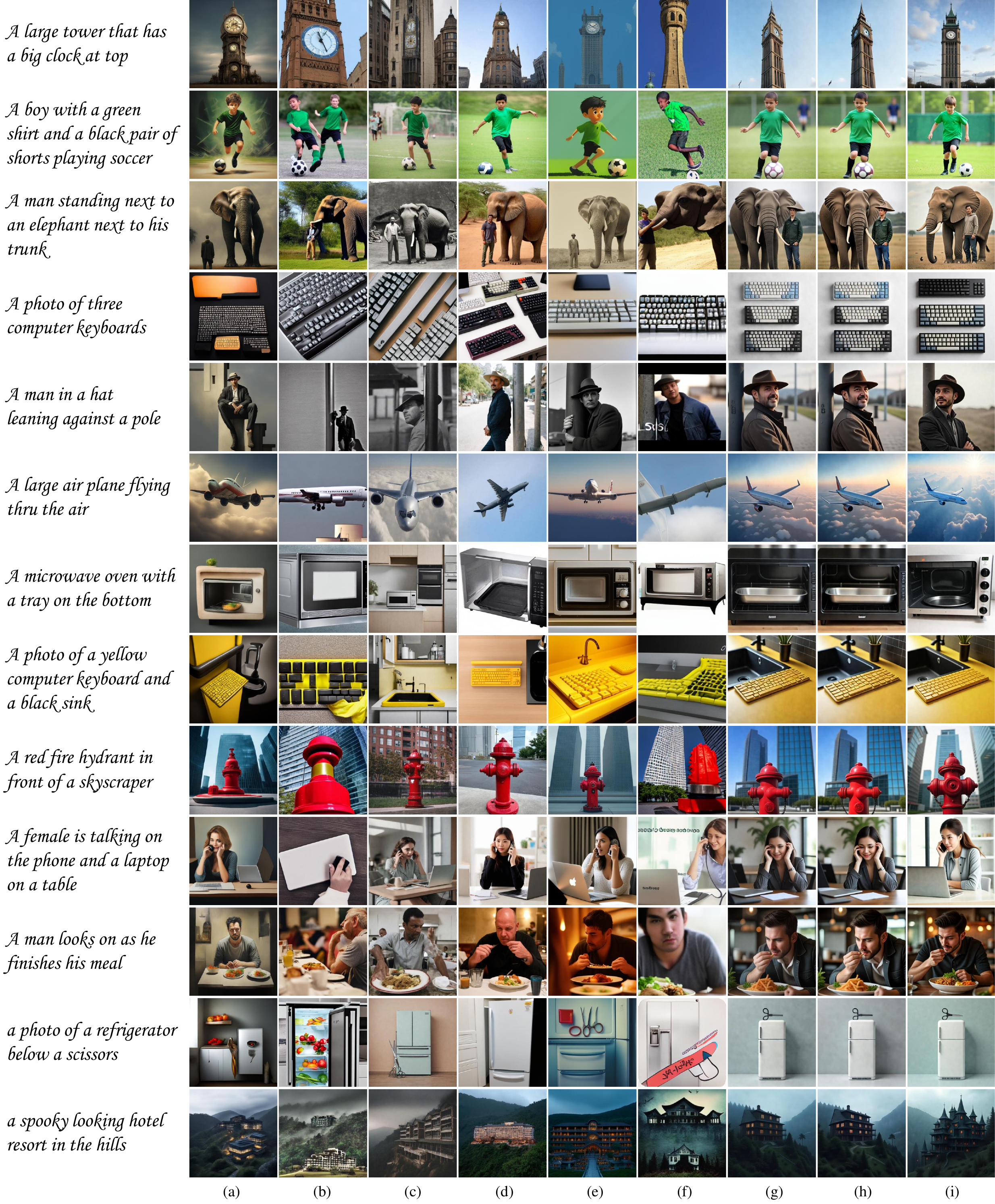}
  \caption{More examples on text-conditioned generation of different methods. From left to right, (a)HART, (b)SD1.5, (c)SDXL, (d)SD3, (e)Show-o, (f)LlamaGen, (g)LaytonGen-H, (h)LaytonGen-H* and (i)LaytonGen-T*.}
  \label{fig:generation_appendix}
\end{figure*}

We present extended comparative results for image reconstruction in Fig. \ref{fig:reconstruction_appendix} and additional text-conditioned image generation outcomes in Fig. \ref{fig:generation_appendix}.

\end{document}